\documentclass[conference]{IEEEtran}
\IEEEoverridecommandlockouts

\usepackage{cite}
\usepackage[utf8]{inputenc}
\usepackage{graphicx}
\usepackage{listings}
\usepackage[dvipsnames]{xcolor}
\usepackage{float}
\usepackage{booktabs}
\usepackage{algorithmic, algorithm, }
\usepackage{threeparttable}
\usepackage{amsmath, amsfonts, amsthm, amssymb, mathtools, xparse, stackengine}
\usepackage{footnote}
\makesavenoteenv{threeparttable}
\usepackage{tablefootnote}
\usepackage{algorithmic, algorithm}
\stackMath
\usepackage{graphicx,float}
\usepackage{subfigure}
\usepackage{nth}
\begin{document}
\title{Off-Policy Correction for Deep Deterministic Policy Gradient Algorithms via Batch Prioritized Experience Replay}

\author{
    \IEEEauthorblockN{Dogan C. Cicek\IEEEauthorrefmark{1}, Enes Duran\IEEEauthorrefmark{1}, Baturay Saglam\IEEEauthorrefmark{1}, Furkan B. Mutlu\IEEEauthorrefmark{1} and Suleyman S. Kozat\IEEEauthorrefmark{1}\IEEEauthorrefmark{2}} \\
    \IEEEauthorblockA{\IEEEauthorrefmark{1}Department of Electrical and Electronics Engineering, Bilkent University, Ankara, Turkey}
    \IEEEauthorblockA{\IEEEauthorrefmark{2}\IEEEmembership{Senior~Member,~IEEE}}
        \IEEEauthorblockA{\tt\footnotesize\{cicek, enesd, baturay, burak.mutlu, kozat\}@ee.bilkent.edu.tr}}
    
\maketitle

\begin{abstract}

The experience replay mechanism allows agents to use the experiences multiple times. In prior works, the sampling probability of the transitions was adjusted according to their importance. Reassigning sampling probabilities for every transition in the replay buffer after each iteration is highly inefficient. Therefore, experience replay prioritization algorithms recalculate the significance of a transition when the corresponding transition is sampled to gain computational efficiency. However, the importance level of the transitions changes dynamically as the policy and the value function of the agent are updated. In addition, experience replay stores the transitions are generated by the previous policies of the agent that may significantly deviate from the most recent policy of the agent. Higher deviation from the most recent policy of the agent leads to more off-policy updates, which is detrimental for the agent. In this paper, we develop a novel algorithm, Batch Prioritizing Experience Replay via KL Divergence (KLPER), which prioritizes batch of transitions rather than directly prioritizing each transition. Moreover, to reduce the off-policyness of the updates, our algorithm selects one batch among a certain number of batches and forces the agent to learn through the batch that is most likely generated by the most recent policy of the agent. We combine our algorithm with Deep Deterministic Policy Gradient and Twin Delayed Deep Deterministic Policy Gradient and evaluate it on various continuous control tasks. KLPER provides promising improvements for deep deterministic continuous control algorithms in terms of sample efficiency, final performance, and stability of the policy during the training.

\end{abstract}

\begin{IEEEkeywords}
deep reinforcement learning, experience replay, prioritized sampling, continuous control, off-policy learning
\end{IEEEkeywords}

\IEEEpeerreviewmaketitle

\section{Introduction}

Deep Reinforcement Learning techniques have shown notable success on tasks that require sequential decision making. Deep Reinforcement learning agents reach the superhuman-level performance on ATARI Games \cite{mnih2013playing}, continuous control tasks \cite{lillicrap2019continuous}, board games \cite{SilverHuangEtAl16nature}, and real-time strategy games \cite{2019Natur.575..350V}. Coupling Reinforcement Learning with Deep Learning enables the agent to learn a parameterized policy and converge to a nearly optimal policy without visiting each state-action pair \cite{Sutton1998}. On the other hand, feeding the neural network that generates the policy of the agent by temporally correlated inputs violates the i.i.d assumption of the stochastic gradient-based optimization algorithms. Experience Replay tackles the given problem and breaks the temporal correlation by stacking the transitions to a replay buffer, then picking mini-batches among them randomly \cite{10.1007/BF00992699}. Due to this process, the agent learns from the transitions that are collected from various states of the state space of the task. The works show that the utilization of experience replay provides improvements to the agent in terms of sample efficiency and the stability of the policy \cite{mnih2013playing}, \cite{mnih2015human}, \cite{DBLP:journals/corr/AndrychowiczWRS17}, \cite{DBLP:journals/corr/HasseltGS15}, \cite{DBLP:journals/corr/WangBHMMKF16}.

Vanilla Experience Replay (Vanilla ER) algorithm samples transitions from the replay buffer randomly. By uniformly sampling the transitions, the algorithm assumes that the importance of each transition is equal to each other. However, the study shows that the strategy on how the agent’s experiences are used during the training drastically affects the performance of the agent \cite{JMLR:v19:17-131}. Several algorithms are suggested on how experiences should be replayed by assigning sampling probabilities and yield promising results \cite{zhang2018deeper}, \cite{schaul2016prioritized}, \cite{oh2021learning}, \cite{Sun_Zhou_Li_2020}, \cite{zha2019experience}.

In this paper, we introduce a novel experience replay prioritization method, Batch Prioritized Experience Replay via KL Divergence, KLPER. We approach the experience replay prioritization problem by prioritizing the sampled batches of transitions rather than assigning sampling probabilities to the transitions. The main drawback of prioritizing the transitions is that the importance of a transition can significantly change until the transition is sampled again. Therefore, the sampling probabilities of the transitions may not be proportional to their actual importance. In addition, as the policy of the agent changes, the replay buffer contains more off-policy transitions. It has shown that more off-policy updates induce divergence and negatively affect the performance of the agent \cite{zhang2018deeper}, \cite{vanhasselt2018deep}. Hence, our algorithm forces the agent to learn through the batch of transitions that are more likely generated by the policy of the agent. We assume that each batch has a policy, Batch Generating Policy, that generalizes the past policies of the agent that collected the transitions in the batch. Then, we define the Batch Generating Policy for each batch with respect to the most recent policy of the agent that. We use the KL Divergence between Batch Generating Policy and a multivariate Gaussian distribution with a mean of 0 as a proxy to measure the deviation between the Batch Generating Policy and the most recent policy of the agent. 

We evaluate KLPER by coupling it with the Deep Deterministic Policy Gradient and the Twin Delayed Deep Deterministic Policy Gradient algorithms. We compare our algorithm with Prioritized Experience Replay and Vanilla ER algorithms, on OpenAi Gym’s continuous control tasks \cite{todorov_erez_tassa_2012}. 

The main contributions of this paper are summarized as follows:

\begin{itemize}
\item Prioritize one batch among certain number of batches that sampled from the replay buffer at each iteration.
\item Define Batch Generating Policy with respect to the most recent policy of the agent to obtain the most likely policy that generates the given batch of transitions.
\item Develop KLPER, to enable the agent learn through more on-policy updates. KLPER uses KL Divergence between Batch Generating Policy and the most recent policy of the agent to prioritize batches of transitions.
\item Demonstrate KLPER on 6 different continuous control tasks. Results yield that our algorithm brings significant improvements on particular tasks in terms of final performance and sample efficiency.
\end{itemize}

\section{Background}

In this section, we briefly introduce the reinforcement learning framework. We summarize two off-policy continuous control algorithms that we couple with the KLPER. Then, we cover various experience replay prioritization methods.

\subsection{Reinforcement Learning}

In reinforcement learning, the agent tries to find a policy that maximizes the cumulative reward while interacting with the environment. Reinforcement Learning tasks can be formulated as Markov Decision Processes (MDP). At each discrete time step $t$, the agent observes a state $s_t \in \mathcal{S}$ from the environment, then the agent selects an action $a_t \in \mathcal{A}$, according to its policy $a_t \sim \pi(a|s_t)$. After an action has been selected, the agent receives a reward and the next state, $s_t' \in \mathcal{S}$ with respect to the learning environment dynamics, $P(s', r|s, a)$. The combination of the four elements, $(s, a, r, s')$, forms a transition, which is appended to a replay buffer that stores the agent’s experiences. The main goal of an agent is maximizing its return, discounted cumulative reward, which is defined as $G_{t} = \sum_{i = t}^{T}\gamma^{i - t}r(s_{i}, a_{i})$, where $\gamma$ is the discount factor.

\subsection{Deep Deterministic Policy Gradient}

Deep Deterministic Policy Gradient (DDPG) is an Off-Policy Deep Reinforcement Learning algorithm that produces deterministic actions on continuous action space. DDPG is an extension of the Deterministic Policy Gradient algorithm \cite{dpg} which includes function approximation \cite{lillicrap2019continuous}. DDPG consists of two nested network architectures that output the policy of the agent, i.e., Actor, and the estimate values of the state-action pairs, i.e., Critic. The Actor network generates a deterministic action with respect to the state, $a = \psi(s;\phi)$. The Critic network outputs a value representing the estimated return after taking action at state $s$, $Q(s,a;\theta)$. In the DDPG algorithm, the Actor and the Critic networks are updated sequentially, starting with the Critic. Bootstrapping by directly using the Actor and the Critic networks makes agents prone to divergence \cite{lillicrap2019continuous}. To achieve stability during the training and avoid a divergent policy, target networks, which drastically improved the performance of the DQN algorithm, are used on DDPG \cite{mnih2015human}. In the DDPG algorithm, there are two networks called the Actor Target and the Critic Target networks which are initialized identically as the Actor and the Critic Networks. Bootstrapping is performed on the Actor Target and the Critic Target networks and their parameters are updated softly updated with respect to the Actor and the Critic networks \cite{lillicrap2019continuous}. The Critic network is trained by using the L2 Loss of the one-step Temporal Difference Error as follows:
\begin{equation}
    L = \mathbb{E}_{(s, a, r, s') \sim \mathcal{D}}[(y- Q(s, a; \theta)^{2}],
\end{equation} where $\phi$ and $\theta$ are the parameters of the Actor network and the Critic network, respectively, $y$ is defined as:
\begin{equation}
    y = r + Q'(s', \psi(s';\phi'); \theta'),
\end{equation}
where $\phi'$ and $\theta'$ are the parameters of the Actor target network and the Critic network, respectively. The Actor network is optimized by taking the derivative of the objective function with respect to the Actor Network parameters:
\begin{equation}
    \nabla_{\theta}J(\theta) = \mathbb{E}_{s \sim p_{\pi}}[\nabla_{a}Q(s, a; \theta)|_{a = \psi(s; \phi)}\nabla_{\phi}\psi(s; \phi)],
\end{equation}
Finally, the Target network parameters are softly updated with respect to the Actor and the Critic parameters.
\begin{equation}
    \phi' = \tau\phi + (1 - \tau)\phi', \quad \theta' = \tau\theta + (1 - \tau)\theta',
\end{equation}
where $\tau$ is the parameter that controls the rate of the soft update.

\subsection{Twin Delayed DDPG}

Twin Delayed Deep Deterministic Policy Gradient (TD3) is an extended version of the DDPG algorithm that significantly improves the performance of its predecessor DDPG \cite{fujimoto2018addressing}. TD3 remedies the problem of overestimation in state-action pairs of the DDPG algorithm. TD3 includes two Critic networks. The minimum Q value that is produced by the Critic Target Networks is set as the target value. In this procedure, called as Clipped Double Q-learning, the target value is calculated as follows:
\begin{equation}
    y = r + \gamma \underset{i=1, 2}{\mathrm{min}}Q'_{i}(s', \psi'(s'; \phi'); \theta'_{i}),
\end{equation}
In addition to using two Critic networks and taking the minimum value generated by these two networks, TD3 makes several modifications on DDPG. TD3 includes Delayed policy update and target policy smoothing, which enables TD3 to outperform DDPG on continuous control tasks \cite{fujimoto2018addressing}.

\subsection{Experience Replay Methods}

The most primitive version of the Experience Replay mechanism is sampling transitions from the replay buffer uniformly. The idea that some transitions might be more useful or more adversarial than the other ones led to the emergence of new sampling methods \cite{zhang2018deeper}.

One of the most well-known techniques is Prioritized Experience Replay (PER) \cite{schaul2016prioritized}. PER increases the sampling probabilities of the transitions that yield more unexpected outcomes for the agent. However, the unexpectedness measure of a transition is not directly reachable. Therefore, PER uses Temporal Difference Error as a proxy to quantify the importance of a transition by assuming a positive correlation between Temporal Difference Error and the unexpectedness of the transition. Temporal Difference Error can be defined as follows:
\begin{equation}
    |\delta|=\mid r+\gamma Q\left(s^{\prime}, a^{\prime} ; \theta^{\prime}\right)-Q(s, a ; \theta)|,
\end{equation}
where $\theta$ and $\theta'$ represent the value and target value network parameters, respectively.

Sharpness of the prioritization can be softened by adjusting the $\alpha$ parameter that combines PER with uniform sampling. Then, the sampling probability of a transition becomes:
\begin{equation}
    P(i)=\frac{p_{i}^{\alpha}}{\sum_{k} p_{k}^{\alpha}},
\end{equation}
Attentive Experience Replay (AER) is an experience replay algorithm that prioritizes transitions by using the state information \cite{Sun_Zhou_Li_2020}. AER assumes that states visited by the past policies are not useful for the training process of the agent. These states should be less frequently visited when the agent has a more stable policy. Then, AER feeds the agent with the frequently visited states. The algorithm aims to optimize the agent by using transitions collected by the recent policies of the agent.

We emphasize that the experience replay prioritization can be defined as a learning task. Then, an expert that has a dynamic policy could learn how to optimize the learning progress of the agent and provide an optimal experience replay strategy for the agent. Experience Replay Optimization and Neural Experience Replay Sampler are two approaches that handle the experience replay prioritization by adding a parameterized replay policy \cite{zha2019experience}, \cite{oh2021learning}. The replay policy component of these methods assigns scores to each transition given the extracted information from the transition such as state, action, the reward of the transition, temporal-difference error, and timestep when the transition is generated. In the following section, we give more details on how we construct our algorithm.

\begin{figure*}[htbp]
	\centering
	
	\subfigure[LunarLanderContinuous-v2]{
		\includegraphics[width=2.35in, keepaspectratio]{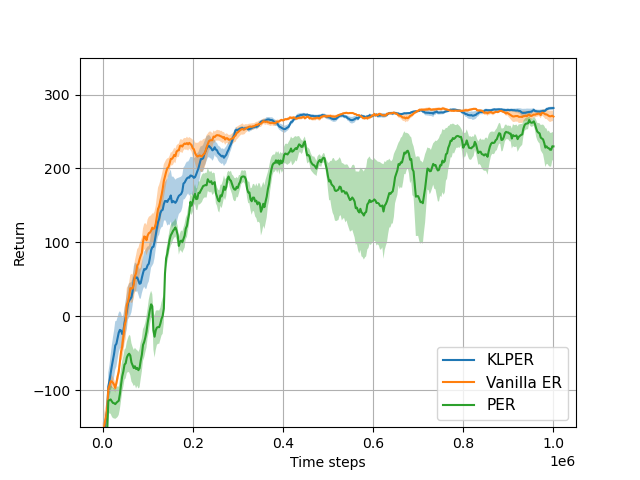}
	}
	\hspace{-0.40in}
	\subfigure[Hopper-v2]{
		\includegraphics[width=2.35in, keepaspectratio]{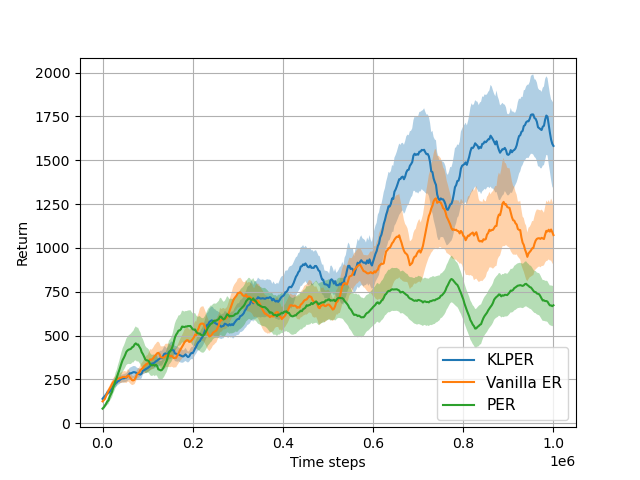}
	}
	\hspace{-0.40in}
	\subfigure[InvertedPendulum-v2]{
		\includegraphics[width=2.35in, keepaspectratio]{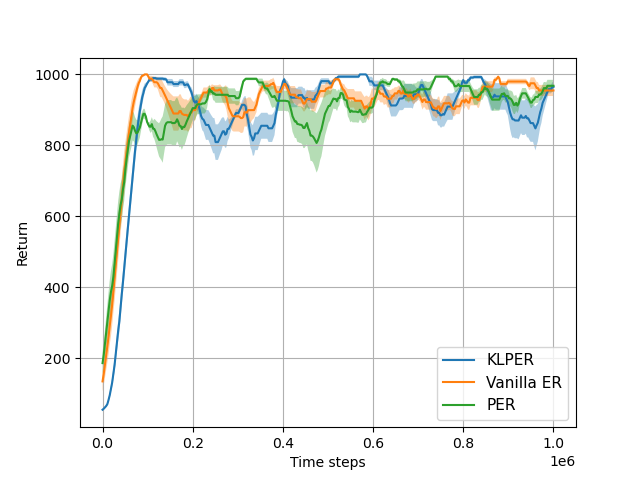}
	}
	\hspace{-0.40in}
	\subfigure[Reacher-v2]{
		\includegraphics[width=2.35in, keepaspectratio]{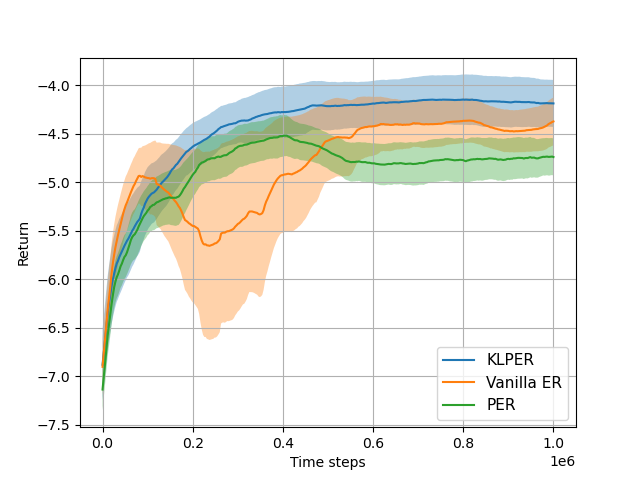}
	}
	\hspace{-0.40in}
	\subfigure[Walker2d-v2]{
		\includegraphics[width=2.35in, keepaspectratio]{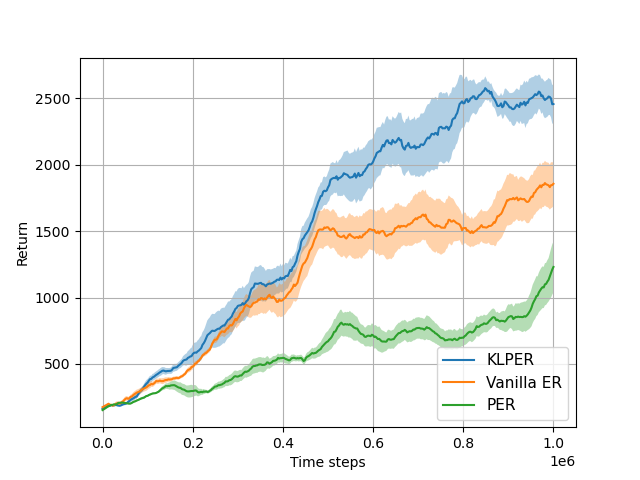}
	}
	\hspace{-0.40in}
	\subfigure[HalfCheetah-v2]{
		\includegraphics[width=2.35in, keepaspectratio]{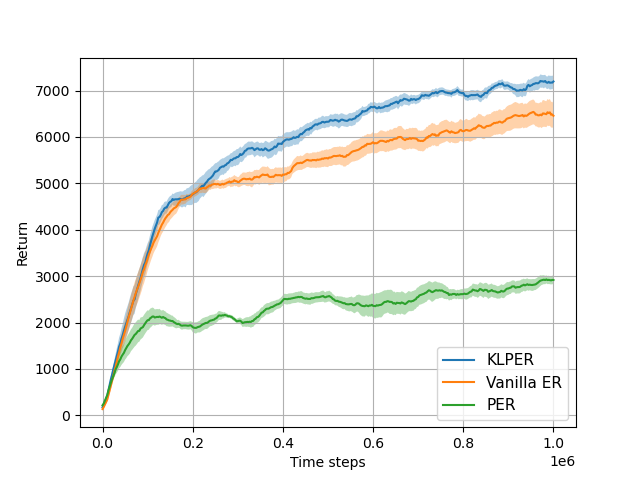}
	}
	
	\caption{Learning Curves of the experience replay methods, KLPER, PER and Vanilla ER on 6 different OpenAI Gym continuous control tasks. The algorithms are coupled with the DDPG. Cumulative reward curves are smoothed for visual clarity. The shaded  regions represents half a standard deviation over five trials.} 
\end{figure*}

\begin{figure*}[htbp]
	\centering
	\subfigure[LunarLanderContinuous-v2]{	\includegraphics[width=2.35in, keepaspectratio]{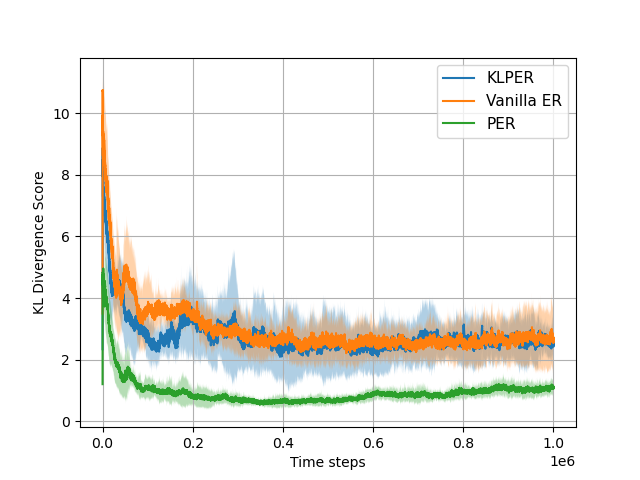}
	}
	\hspace{-0.40in}
	\subfigure[Hopper-v2]{
		\includegraphics[width=2.35in, keepaspectratio]{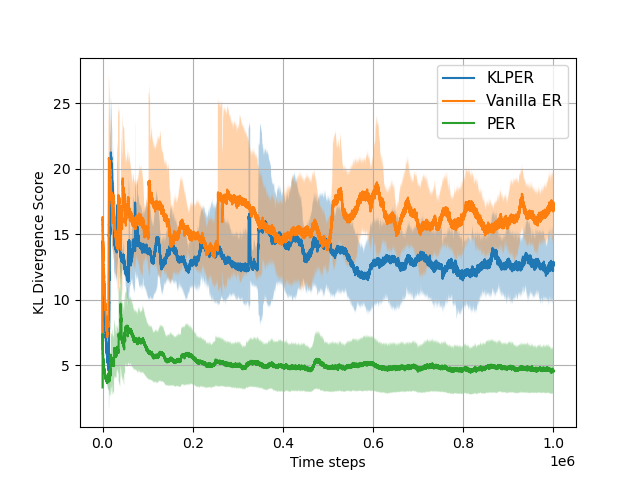}
	}
	\hspace{-0.40in}
	\subfigure[InvertedPendulum-v2]{
		\includegraphics[width=2.35in, keepaspectratio]{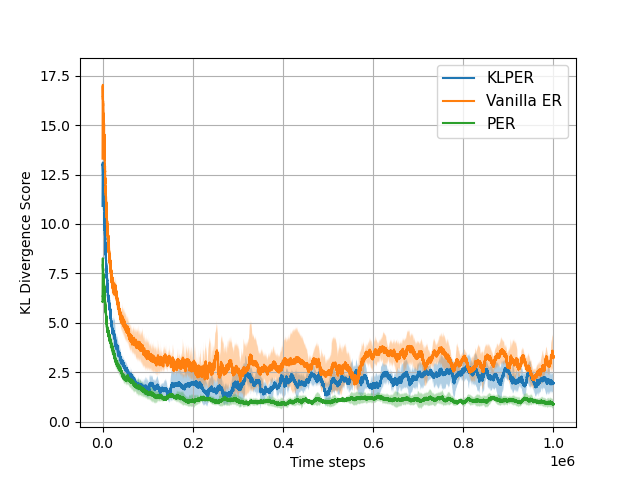}
	}
	\hspace{-0.40in}
	\subfigure[Reacher-v2]{
		\includegraphics[width=2.35in, keepaspectratio]{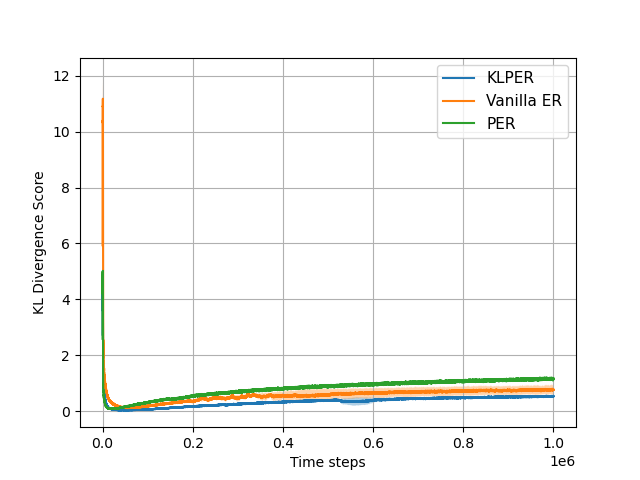}
	}
	\hspace{-0.40in}
	\subfigure[Walker2d-v2]{
		\includegraphics[width=2.35in, keepaspectratio]{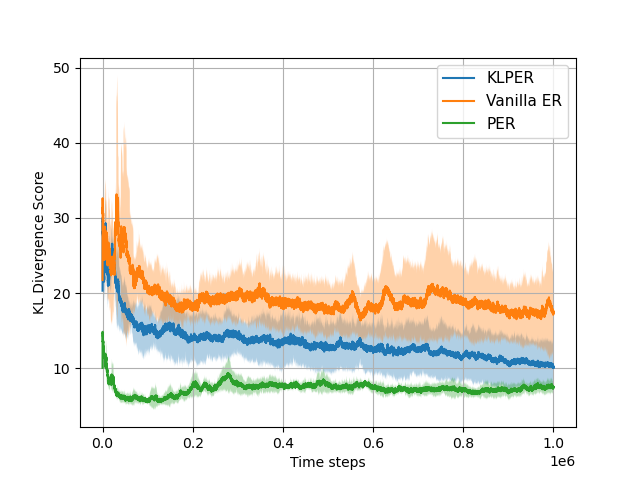}
	}
	\hspace{-0.40in}
	\subfigure[HalfCheetah-v2]{
		\includegraphics[width=2.35in, keepaspectratio]{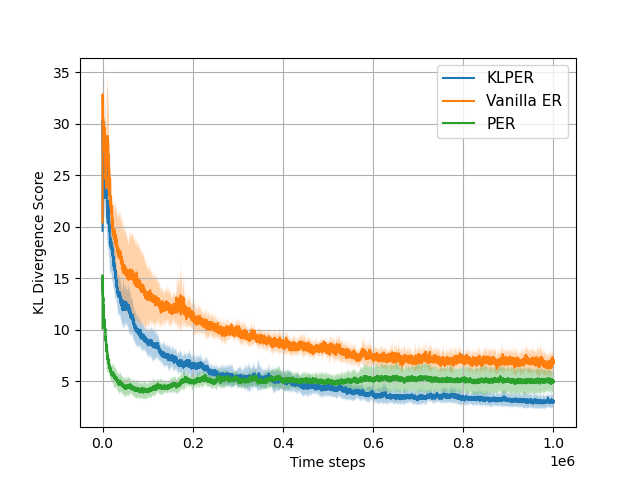}
	}
	
	\caption{KL Divergence scores that are yielded by Batch Generating Policy and multivariate Gaussian distribution with mean 0 and covariance 0.1$\mathbb{I}$ for each algorithm that coupled with the DDPG. Curves are smoothed for visual clarity.} 
\end{figure*}

\section{KL Experience Replay}

In this section, we mention the problems that KLPER aims to tackle. We define the Batch Generating Policy, one of the critical components of the algorithm. Then, we give more details about the batch selection process among candidate batches.

\subsection{Motivation}

Increasing sampling probability of the specific transitions over the other ones may lead to undesirable updates on the Actor and the Critic networks. The likelihood of having these unwanted updates is proportional to the heaviness of the prioritization since heavy prioritization causes more Off-Policy Updates \cite{vanhasselt2018deep}.  

The sampling probability of a transition reflects the importance of the transition. Each transition’s contribution to the learning process depends on the policy of the agent and the value network that the agent uses. To quantify the importance of a transition, one measure should be defined. For instance, PER uses Temporal Difference error for that purpose \cite{schaul2016prioritized}. Furthermore, the importance of the transitions changes during the training since the policy of the agent or the value network that the agent uses is updated after each iteration. Then, to properly arrange the importance of the samples in the buffer, one should span the whole replay buffer and recalculate the sampling probabilities, which is infeasible in terms of computation after some point since the number of samples in the buffer increases rapidly. PER uses a practical solution for the given problem, recalculates a transitions sampling probability when it is sampled \cite{schaul2016prioritized}. In that case, the expected sampling period of a prioritized transition is calculated as follows:
\begin{equation}
    T_s = \frac{1}{P_i b}
\end{equation}
where $P_i$ is the $i$th transition's sampling probability and $b$ is the mini-batch size. This procedure assumes that the importance of a transition remains the same until it is sampled again. However, a desirable transition may become indifferent or even adversarial at the latter stages of the training for the agent and vice versa \cite{zhang2018deeper}. Due to the aforementioned issues, an algorithm that prioritizes transitions may lead to an undesirable training process. Then, the Vanilla ER algorithm that samples transitions may outperform a method that prioritizes the samples in the replay buffer \cite{vanhasselt2018deep}.

In reinforcement learning, the deadly triad is defined as the combination of function approximation, bootstrapping, and off-policy learning. An algorithm that includes these three properties may suffer from unbounded value estimates, which deteriorates the learning process of the agent \cite{vanhasselt2018deep}. Function Approximation is the most indispensable component of reinforcement learning among the properties of the deadly triad because when state and action spaces of the reinforcement learning task are huge, then visiting all state-action pairs becomes infeasible, especially in the continuous domain. One may use Monte Carlo learning instead of Temporal Difference learning, then discard the Bootstrapping. However, Monte Carlo learning requires long trajectories that end with a terminal state. Tasks that have no termination conditions cannot be properly solved by Monte Carlo learning. If the On-Policy learning is chosen over Off-Policy learning, the agent cannot use the experiences generated from its past policies. Then, the heavily correlated transitions negatively affect the neural network training \cite{vanhasselt2018deep}. The final component of the deadly triad, Off-Policy learning, can be softened by changing the sampling probabilities of the transitions and the performance of the modified algorithm performs better on the learning tasks \cite{vanhasselt2018deep}.

In this paper, we introduce a novel experience replay prioritization algorithm, KLPER, which reduces the off-policyness of the updates at each iteration for Deep Deterministic Policy Gradient algorithms. Our approach selects one batch among candidate batches rather than assigning sampling probabilities to the transitions in the replay buffer.

\subsection{Batch Generating Policy}

Behavior policy is defined as a mixture of an exploration noise and the target policy of the agent, which is parameterized by the Actor network, for DDPG and TD3 algorithms. The replay buffer of an agent includes transitions gathered by the past policies of the agent. We attempt to find the most likely policy that generates the sampled batch of transitions with respect to the most recent policy of the agent. We call that policy as Batch Generating Policy and denote it as $\omega$. We remark that the policy used for generating each transition becomes intractable after the transition is stored due to the exploration noise term of the behavior policy. Thus, we assume that the Batch Generating Policy is stochastic. We elaborate on how we build the Batch Generating Policy for the remaining part of this subsection.

Feedforwarding the states in the batch of transitions, $\boldsymbol{S}^{b \times m}$, to the Actor network yields the actions, $\boldsymbol{\hat{A}}^{b\times l}$ that the agent act in these states respect to its most recent policy:
\begin{equation}
    \boldsymbol{\hat{A}}^{b\times l} = \psi(\boldsymbol{S}^{b \times m}; \phi),
\end{equation}
where $b$ is the mini-batch size, $\psi$ is the Actor network, $l$ and $m$ are the number of dimensions that action and state space have, respectively. The difference between actions in the batch and actions that would be taken by the agent’s most recent policy, $\boldsymbol{\dot{A}}^{b\times l}$, represents the deviation between the current policy of the agent and previous policies of the agent that were used to generate experiences.
\begin{equation}
    \boldsymbol{\dot{A}}^{b\times l} := \boldsymbol{\hat{A}}^{b\times l} - \boldsymbol{A}^{b\times l},
\end{equation}
where $\boldsymbol{A}^{b\times l}$ is the actions stored in the transitions of the sampled batch. The exploration noise choice affects the behavior policy of the agent. The works show that using Gaussian noise as the exploration noise rather than Ornstein-Uhlenbeck noise \cite{UhleOrns30} does not decrease the performance of the DDPG algorithm \cite{fujimoto2018addressing}, \cite{fujimoto2019offpolicy}. Hence, We choose Gaussian noise,as the exploration noise for both algorithms. 

We remark that Batch Generating Policy is stochastic, and we defined it as a probability distribution. The mean of the distribution can be formulated as the difference between the actions of the sampled transitions and the actions that the agent would produce with respect to states of the corresponding transition:
\begin{equation}
    \mu_{\omega}^{1 \times l} = \frac{1}{b} \sum_{i \in b} \boldsymbol{\dot{A}}^{b\times l}_{ij},
\end{equation}
where $i$ is the $i$th element of the batch and $j$ is the $j$th dimension of the action space. We define the covariance matrix of the distribution as follows:
\begin{equation}
    \mathrm{\Large\Sigma}_{\omega}^{l \times l} = \frac{1}{b-1} \sum_{k \in b}  (\boldsymbol{\dot{a}}_k^{1 \times l}-\mu_{\omega}^{1 \times l})^{\top}(\boldsymbol{\dot{a}}_k^{1 \times l}-\mu_{\omega}^{1 \times l}),
\end{equation}
where $\boldsymbol{\Dot{a}}_k^{1 \times l}$ is the $k$th row of the $\boldsymbol{\dot{A}}^{b\times l}$ matrix. We define the shape of the distribution as the multivariate Gaussian by Maximum Entropy Principle. Finally, we obtain Batch Generating Policy:
\begin{equation}
    \omega \sim \mathcal{N}(\mu_{\omega}^{1 \times l}, \mathrm{\Large\Sigma}_{\omega}^{l \times l}).
\end{equation}

\subsection{Choosing Batches with KL Divergence}
In this section, we give more details on how KLPER scores and chooses one batch among candidate batches.

Firstly, at each training timestep $t$, KLPER samples $N$ candidate batches before updating the parameters of the Actor and the Critic networks. Then, the algorithm derives Batch Generating Policy for each sampled batch. It uses KL Diverge to measure the similarity between the policy of the agent and the Batch Generating Policy. The Actor networks for the DDPG and TD3 algorithms yield deterministic policies. However, we remark that the transitions are generated by following the behavior policy of the agent. Therefore, we define the target distribution for the KL Divergence as a multivariate Gaussian distribution with mean 0 and variance $\sigma \mathbb{I}$. We refer KL score for each batch with $\kappa$ and KLPER calculates the KL Score as follows:
\begin{equation}
    \kappa = \mathrm{D}_{\mathrm{KL}}(\mathcal{N}(\mu_\omega, \Sigma_\omega) \| \mathcal{N}(0, \sigma \mathbb{I})),
\end{equation}
where $\mathbb{I}$ is the identity matrix. KLPER selects the batch that yields the minimum KL score among candidate batches to provide the learning algorithms more on-policy updates at each iteration. We provide KLPER in Algorithm 1.

\begin{figure*}[htbp]
	\centering
	
	\subfigure[LunarLanderContinuous-v2]{
		\includegraphics[width=2.35in, keepaspectratio]{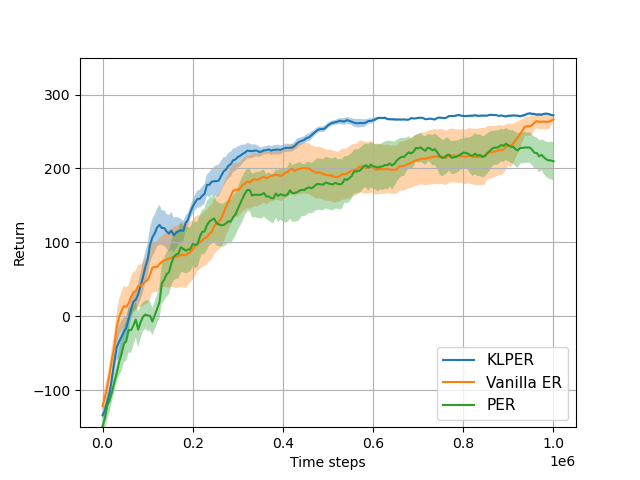}
	}
	\hspace{-0.4in}
	\subfigure[Hopper-v2]{
		\includegraphics[width=2.35in, keepaspectratio]{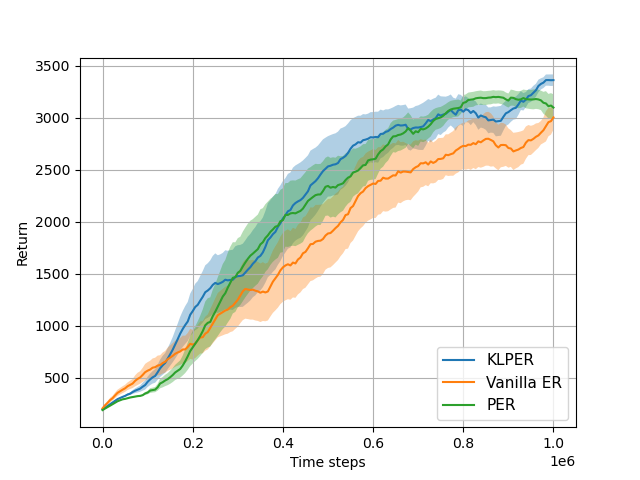}
	}
	\hspace{-0.4in}
	\subfigure[InvertedPendulum-v2]{
		\includegraphics[width=2.35in, keepaspectratio]{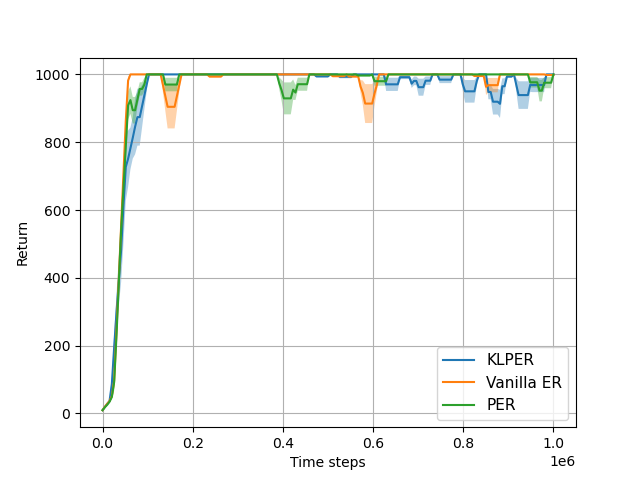}
	}
	\hspace{-0.4in}
	\subfigure[Reacher-v2]{
		\includegraphics[width=2.35in, keepaspectratio]{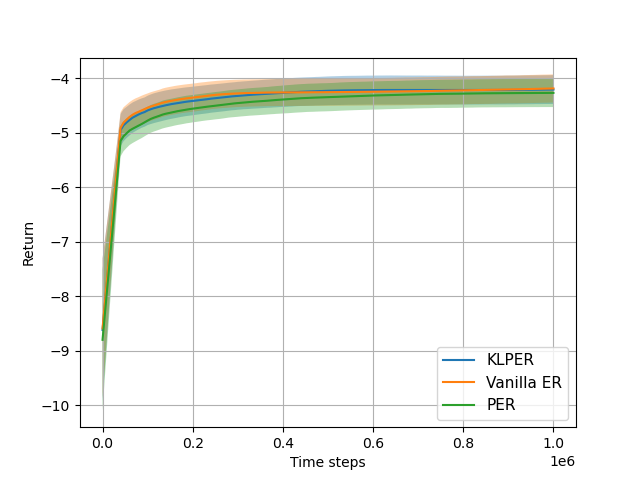}
	}
	\hspace{-0.4in}
	\subfigure[Walker2d-v2]{
		\includegraphics[width=2.35in, keepaspectratio]{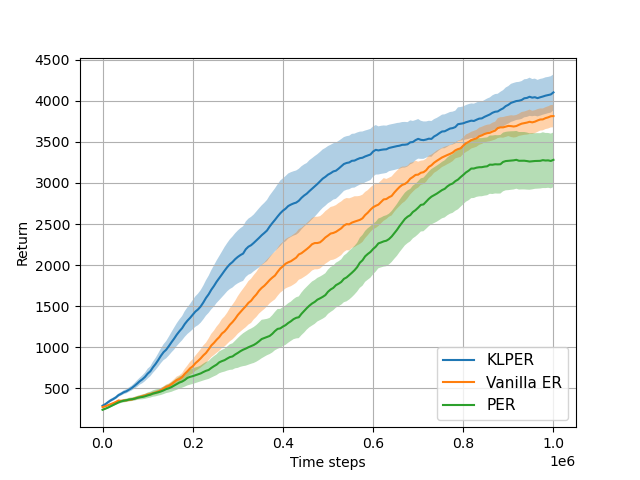}
	}
	\hspace{-0.4in}
	\subfigure[HalfCheetah-v2]{
		\includegraphics[width=2.35in, keepaspectratio]{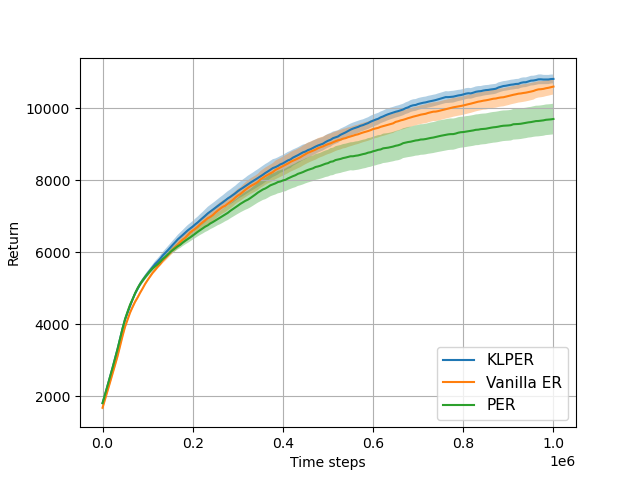}
	}
	
    	\caption{Learning Curves of the experience replay methods, KLPER, PER and Vanilla ER on 6 different OpenAI Gym continuous control tasks. The algorithms are coupled with the TD3. Cumulative reward curves are smoothed for visual clarity. The shaded  regions represents half a standard deviation over five trials.} 
\end{figure*}

\begin{figure*}[htbp]
	\centering
	
	\subfigure[LunarLanderContinuous-v2]{
		\includegraphics[width=2.35in, keepaspectratio]{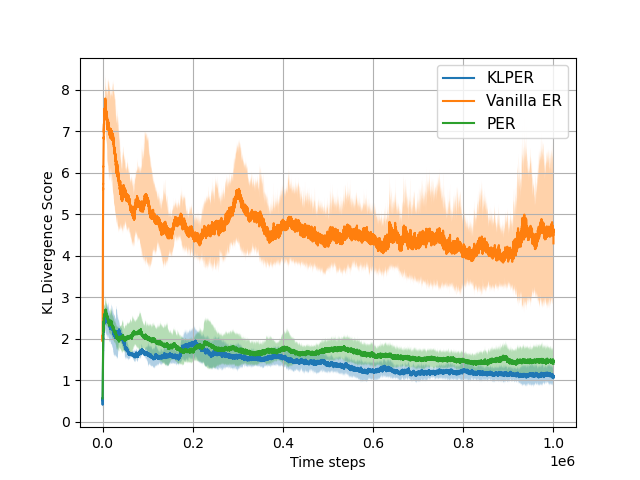}
	}
	\hspace{-0.40in}
	\subfigure[Hopper-v2]{
		\includegraphics[width=2.35in, keepaspectratio]{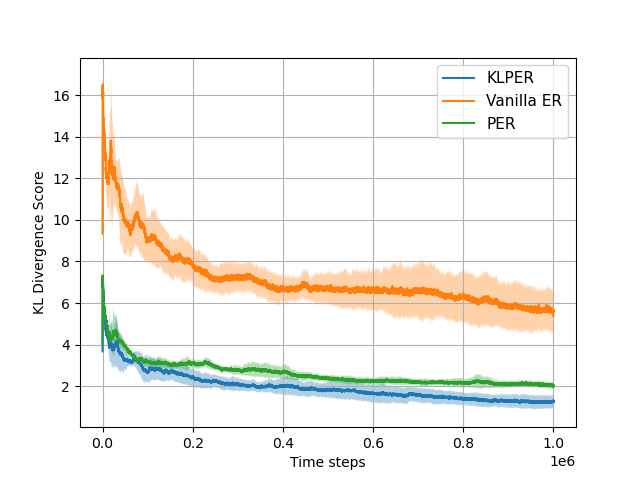}
	}
	\hspace{-0.40in}
	\subfigure[InvertedPendulum-v2]{
		\includegraphics[width=2.35in, keepaspectratio]{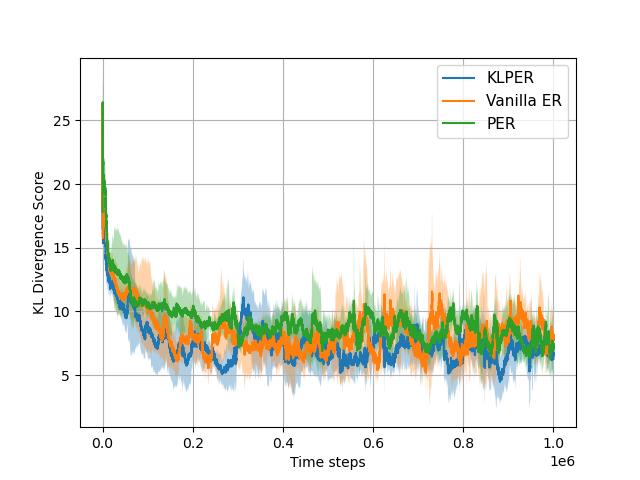}
	}
	\hspace{-0.40in}
	\subfigure[Reacher-v2]{
		\includegraphics[width=2.35in, keepaspectratio]{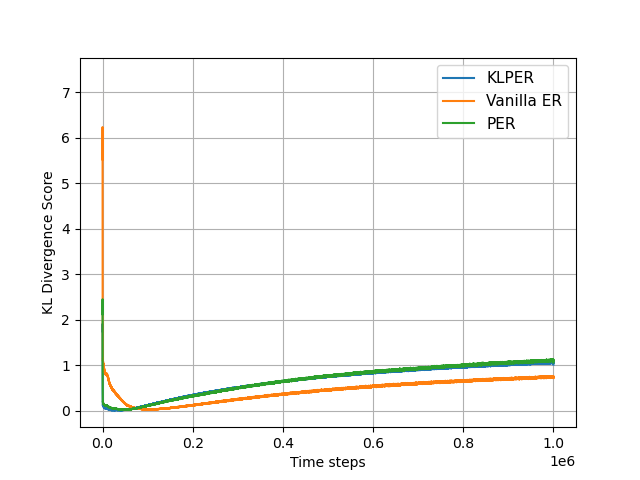}
	}
	\hspace{-0.40in}
	\subfigure[Walker2d-v2]{
		\includegraphics[width=2.35in, keepaspectratio]{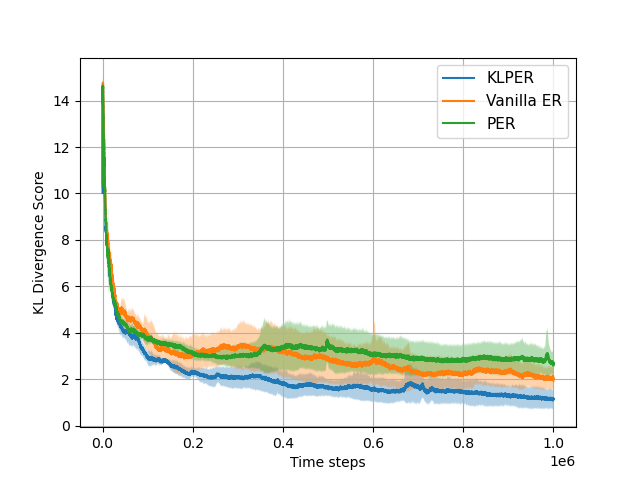}
	}
	\hspace{-0.40in}
	\subfigure[HalfCheetah-v2]{
		\includegraphics[width=2.35in, keepaspectratio]{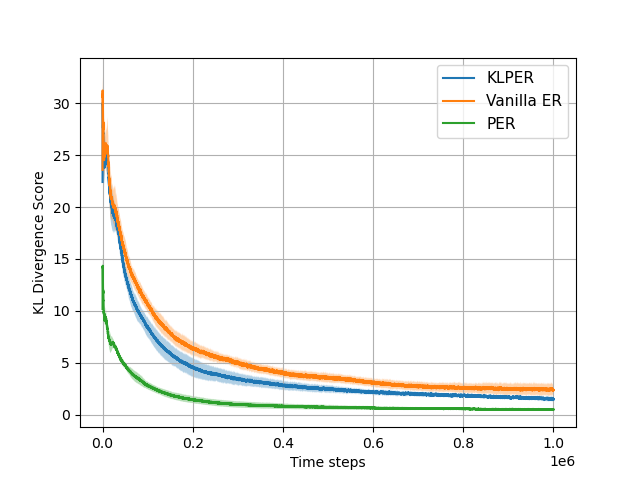}
	}
	
	\caption{KL Divergence scores that are yielded by Batch Generating Policy and multivariate Gaussian distribution with mean 0 and covariance 0.1$\mathbb{I}$ for each algorithm that coupled with the TD3. Curves are smoothed for visual clarity.} 
\end{figure*}


\section{Experiments}
In this section, we elaborate on the implementation details on learning algorithms and hyper-parameters. Then, we provide results and comparisons of the KLPER with Vanilla ER and PER. We inspect the results into two categories: results for DDPG and results for TD3.

\begin{algorithm}[htbp]
    \caption{KLPER}
    \begin{algorithmic}
        \STATE Initialize batch size $b$, replay buffer $\mathcal{D}$
        \STATE Initialize $M$ for delayed policy updates
        \STATE Initialize number of candidate batches $N$
        \STATE Initialize $\sigma$ for the target distribution 
        \FOR{$t = 1$ \textbf{to} $T$}
            \STATE Observe $s_t$ Choose action $a_t \sim \pi_{\phi}(a|s_t) + \epsilon$
            \STATE Observe reward $r_t$ and new state $s_t'$
            \STATE Store transition $(s_t, a_t, r_t, s_t')$ in $\mathcal{D}$
            \STATE Sample $N$ batches from $\mathcal{D}$
            \FOR{$n = 1$ \textbf{to} $N$}
                \STATE Calculate $\kappa_n$ for the batch (Eq. 14)
            \ENDFOR
            \STATE Select the batch that yields minimum $\kappa$
            \STATE Update the weights of the Critic network
            \IF{$k$ mod $M$}
                \STATE Update the weights of the Actor network
                \STATE Update the weights of the Target networks
            \ENDIF
        \ENDFOR
    \end{algorithmic}
    \label{alg:klper}
\end{algorithm}

\subsection{Implementation Details}

We note that our algorithm compared with Vanilla ER and PER. To make a fair comparison, we have run the algorithms by using 5 different random seeds. For each experience replay method, we use the same architecture and hyperparameters for DDPG and TD3 algorithms. Also, we take the network architectures from the original implementations of the DDPG and TD3 algorithms \cite{lillicrap2019continuous}, \cite{fujimoto2018addressing}. 

DDPG has two hidden layer neural networks for both the Actor and the Critic networks, which include 400 and 300 neurons, respectively. For TD3, we use two hidden layer networks that have both 256 neurons, and the network architecture for Actor and Critic are the same. We use Adam as the optimizer of the networks for both DDPG and TD3 \cite{kingma2017adam}. The learning rates of the Actor and the Critic networks are $1 \times 10^{-4}$ and $3 \times 10^{-4}$, respectively, for the DDPG. The learning rate of both Actor and Critic networks for the TD3 algorithm was $1 \times 10^{-3}$. We set the mini-batch size to 64 and 256 for DDPG and TD3. We observe that the DDPG algorithm sticks at the local optima when the number of exploration steps is insufficient. Therefore, we fill the replay buffer with 10000 transitions gathered by the agent while acting randomly for the DDPG algorithm for each experience replay method. The number of exploration steps set to 25000 for TD3 algorithm.

We choose $N$, an adjustable parameter for the algorithm, as 4 for DDPG and 8 as TD3. We choose $\sigma$ as $0.1$ for DDPG to couple exploration noise that is used for the DDPG algorithm and target distribution of the KL Divergence. For TD3, we assign $\sigma$ to $0.2$ since TD3 uses SARSA like updates on bootstrapping step of the algorithm by adding a noise term to the action produced by the Actor Target network. This noise component is sampled from a Gaussian Distribution with a $0.2$ variance.

We use Proportional PER for the comparison. The alpha and the beta parameters set to 0.6 and 0.4 for Proportional PER \cite{schaul2016prioritized}.

For the comparison of the methods, we run the algorithms on 6 different MuJoCo tasks as the learning environments , which vary in terms of state and action spaces. We evaluate the performances of the algorithms for every 5000 timesteps. During the evaluation episodes, agents perform by using their most recent policy 5 times. We assign average cumulative reward over 5 evaluation episodes as the evaluation score of an agent.

\begin{table*}[htbp]
	\caption{State and Action Space dimensions of the environments in MuJoCo control suite}
	\begin{center}
	\begin{threeparttable}
		\begin{tabular}{c|c c c c c c c c}
			\toprule 
			\textbf{Environment} & LunarLander-v2\tnote{1} & Hopper-v2 & InvPend-v2\tnote{2} & Reacher-v2 & Walker2d-v2 & HalfCheetah-v2 \\ \hline
			\textbf{State Dimension} & 8 & 11 & 4 & 11 & 17 & 17 \\ \hline
			\textbf{Action Dimension} & 2 & 3 & 1 & 2 & 6 & 6 \\
			\bottomrule 
		\end{tabular}
	    \begin{tablenotes}
            \item[1] Abbreviation for LunarLanderContinuous-v2
            \item[2] Abbreviation for InvertedPendulum-v2
        \end{tablenotes}
	\end{threeparttable}
	\end{center}
	\label{env_state_action}
\end{table*}

\subsection{Results for DDPG}

In this section, we combine KLPER, PER and Vanilla ER with the DDPG algorithm. In Fig. 1, we propose the learning curves.

Fig. 1 shows that our method outperforms Vanilla ER and PER methods for four of the six learning environments in terms of the model’s final performance and sample efficiency. Cumulative Rewards of the DDPG Agents that use KLPER as the experience replay method almost monotonically increase during the training process for most of the learning environments. For the LunarLanderContinuous-v2, Vanilla ER and KLPER converge to a nearly optimal policy at the early stages of the training. Results on InvertedPendulum-v2 suggest that there is no significant difference among the methods. 

We compare each Batch Generating Policy’s deviation from the most recent policy of the agent for each method and provide the results in Fig. 2. As an interesting finding, even though KLPER samples 4 batches and selects the batch that has the least KL score, mostly batches that selected by PER have lower KL scores during the training process. As an explanation for that, we zoom in on the dynamics of the experience replay. The agent fills its replay buffer after the exploration steps by using its behavior policy which is a stochastic variant of its target policy. If the policy of the agent changes considerably after each policy update, then more off policy transitions would be stored to the replay buffer. We conjecture that KLPER leads more prominent policy changes during the training. Consequently, the sampled batches' KL scores are relatively high when compared to PER.

\subsection{Results for TD3}
In this section, we combine KLPER, PER, and Vanilla ER with the TD3 algorithm. In Fig. 3, we propose learning curves.

KLPER outperforms the agents that use Vanilla ER and PER in four out of six learning environments in terms of the final performance and sample efficiency. For Reacher-v2, agent performances are almost indistinguishable. For LunarLanderContinuous-v2, the variances of the cumulative rewards collected by the agents that use Vanilla ER and PER are relatively high. On the other hand, all the agents that use KLPER converged to a nearly optimal and robust policy. KLPER provides the same performance as the other methods for the InvertedPendulum-v2 task when coupled with both the learning algorithms, DDPG and TD3. We denote the state and action space dimensions of the learning environments in Table 1. The InvertedPendulum-v2 task has four-dimensional state space and one-dimensional action space. Therefore, we explain that fact as KLPER algorithm may fail to work in low dimensional tasks. However, it works well on high-dimensional action and state spaces.

We investigate KL scores of the sampled batches for each method by following the same procedure as in DDPG and provide the results in Fig. 4. It is surprising that KL Divergence values of batches when PER and KLPER are used are so close to each other in LunarLanderContinuous-v2 and Hopper-v2. However, KLPER agents perform better than PER on both of the environments in terms of the final performance. Prioritizing experience replay leads to relatively smaller policy changes on these learning tasks. For InvertedPendulum-v2, KL score curves are noisy for each method. Then we observe that all the experience replay methods, when combined with TD3, do not ensure a robust policy for InvertedPendulum-v2 task.

\section{Conclusion}
In this paper, we emphasize the drawbacks of prioritizing transitions, and improve the performance of the agent by reducing the off-policyness of the reinforcement learning algorithms. We introduce the KLPER algorithm that makes prioritization on the batch of transitions rather than prioritizing each transition in the replay buffer. We define Batch Generating Policy to quantize the off-policyness level of a batch. Our algorithm enables agents to have more on-policy updates using KL Divergence between Batch Generating Policy and a multivariate Gaussian distribution with a mean of 0. We combine KLPER with Deep Deterministic Policy Gradient algorithms and test it on continuous control tasks. Results show that KLPER brings promising improvements and outperforms Vanilla ER and PER in terms of sample efficiency and final performance in particular continuous control tasks. 

\bibliography{offpol}
\bibliographystyle{ieeetr}

\end{document}